\documentclass[a4paper, 11pt]{article}
\usepackage[top=3cm, bottom=3cm, left = 2cm, right = 2cm]{geometry} 
\usepackage[utf8]{inputenc}
\usepackage{textcomp}
\usepackage{graphicx}
\usepackage{epstopdf}
\usepackage{float} 
\usepackage{amsmath,amssymb}  
\usepackage{bm}  
\usepackage[pdftex,bookmarks,colorlinks,breaklinks]{hyperref}  
\usepackage{memhfixc} 
\usepackage{pdfsync}  
\usepackage{fancyhdr}

\usepackage{cleveref}
\usepackage{xspace}
\usepackage{todonotes}
\usepackage[english]{babel}
\usepackage[backend=biber]{biblatex}
\usepackage{csquotes}

\newcommand{\nbbert}{Nb-BERT-large\xspace}
\newcommand{\norbert}{NorBERT3-large\xspace}
\newcommand{\modernbert}{ModernBERT\xspace}
\newcommand{\knbbert}{Kl-Nb-BERT\xspace}
\newcommand{\knorbert}{Kl-NorBERT3\xspace}
\newcommand{\kmodernbert}{Kl-ModernBERT\xspace}
\newcommand{\medMCQA}{MedMCQA\xspace}
\newcommand{\medMQP}{MedMQP\xspace}
\newcommand{\medner}{Nor-DeID\xspace}
\newcommand{\AISMEC}{AISMEC\xspace}
\newcommand{\Synergi}{Synergi\xspace}
\newcommand{\suite}{KliniskVestBERT\xspace}
\title{
    \suite: BERT Model Specialised to Norwegian Clinical Texts
}
\author{Christian Autenried \\ email: \href{mailto:christian.autenried@helse-vest-ikt.no}{christian.autenried@helse-vest-ikt.no}
    \and Cosimo Persia \\  email: \href{mailto:cosimo.damiano.persia@helse-vest-ikt.no}{cosimo.damiano.persia@helse-vest-ikt.no} }
\date{}

\begin{document}
\maketitle
\begin{abstract}
    The increasing application of Natural Language Processing (NLP) in healthcare demands
    language models specifically attuned to the complexities of clinical language. This
    work introduces \suite, a suite of three BERT-based encoder models pre-trained
    on a substantial corpus of real-world, de-identified Norwegian clinical texts from Helse Vest. 
    We
    continue pretraining existing language models  
    \nbbert, \norbert, and \modernbert
    on our specialized clinical
    dataset. This dataset is based on a representative population of Helse Vest patients. The included documenttypes 
    are carefully curated to encompass a broad clinical spectrum in bokmål and nynorsk including 
    discharge summaries, surgical reports, nursing notes etc. ensuring comprehensive representation
    of the linguistic landscape within Norwegian healthcare settings.  
    Evaluation on three synthtetical
    Norwegian clinical benchmark datasets and two real-world problems demonstrates that each of 
    our clinically specialized
    models consistently outperforms their baseline counterparts, highlighting the significant
    benefit of domain-specific pre-training for NLP tasks within the clinical domain. 

    The project was a joint effort by all Helse Vest entities (Helse Bergen, Helse Fonna, 
    Helse Førde and Helse Stavanger) with DIPS under the project lead of Helse Vest ICT.
\end{abstract}

\section{Introduction}
Large language models (LLMs) have recently demonstrated remarkable capabilities across a diverse range 
of natural language processing tasks, sparking considerable interest in their application within the 
medical domain~\cite{huang2020c,NGOPD2025}. 
The potential benefits are substantial, 
ranging from 
automated classification of reports~\cite{Solti2009bu},
question answering~\cite{Overby2009pa} 
to
clinical decision support to improved patient-physician communication~\cite{persadetc}.
However, realizing this potential is significantly hampered by a critical bottleneck: the availability of 
high-quality, labelled medical data for training and adaptation. Unlike many other NLP applications where 
large, publicly available datasets exist, medical texts are inherently sensitive, governed by stringent 
privacy regulations, and often require extensive de-identification and specialized annotation. 
This data scarcity frequently compels developers, medical practitioners and researchers to rely on pre-trained 
LLMs initially trained on general 
domain corpora text lacking the complex medical vocabulary, and specific knowledge base essential for 
accurate and reliable performance in clinical settings. 
Consequently, while promising, the direct application of these general-purpose models often 
yields suboptimal results when confronted with the complications of medical language and tasks.

The lack of sufficient medical training data represents a fundamental challenge to the widespread 
adoption of LLMs in healthcare. While techniques like data augmentation and synthetic data generation 
offer partial solutions, they often struggle to fully capture the complexity and variability present in 
real-world clinical documentation. Furthermore, models trained solely on publicly available 
medical resources, such as PubMed abstracts\footnote{\url{ttps://pubmed.ncbi.nlm.nih.gov/}} or medical textbooks, may not adequately represent the 
language patterns found in patient records, which are often characterized by less formal 
writing styles, abbreviations, and a greater degree of contextual specificity. 
This discrepancy can lead to issues in tasks requiring a specialised understanding of patient history, 
clinical approximated reasoning, and the subtle cues embedded within unstructured electronic health records.  
A need exists for robust language models specifically adapted to, and trained on, representative clinical data 
while maintaining patient privacy and adhering to ethical guidelines.

To address this critical gap, in this work we present three specialised language models 
to norwegian clinical texts, 
fine-tuned 
specifically on a substantial corpus of $16.2$ million de-identified clinical text sourced from patient records 
within the Vestland healthcare region in Norway. 
This dataset, comprising a diverse range of discharge summaries, surgical reports, nursing notes etc.
offers a unique opportunity to train a model that is attuned to the specific linguistic characteristics and 
contextual nuances of real-world clinical practice.  
We run de-identification algorithms on the dataset
to ensure patient privacy and compliance with Norwegian regulations
prior to model training. The resulting model, built upon a pre-trained model, 
aims to bridge the performance gap between general-purpose LLMs and specialized 
clinical language understanding systems, offering a more accurate and reliable 
tool for a variety of medical applications.

In this study, we demonstrate the performance improvements achievable through fine-tuning a 
language model on clinically sourced data. We evaluated our model's capabilities across 
three distinct medical NLP tasks and two real-world problems: 
\begin{itemize} 
    \item \medMCQA: is a large-scale multiple-choice question answering
        (MCQA) dataset assembled from authentic medical entrance exam questions. 
        The dataset challenges models to select the correct answer(s)
        from a provided set of options.
    \item \medMQP: focuses on the task of semantic
        similarity assessment within patient-generated questions. The dataset requires
        models to distinguish between question pairs with identical intent but differing syntax.
    \item \medner:  is a synthetic dataset of Norwegian discharge summaries originally designed for 
        de-identification benchmarking, but adaptable for Named Entity Recognition (NER) tasks.
    \item \AISMEC~\cite{AISMEC2025}: is a research project with the aim to identify stroke patients in emergency calls 
        (Classification task).
    \item \Synergi~\cite{Synergi2025}: is an innovation project in Helse Bergen with the main goal to
        identify pharmaceuticals in text descriptions of unwanted hospital events (NER).
\end{itemize}
We utilize open benchmark datasets in each domain.  
Our results consistently demonstrate that our clinical-data finetuned model has better performance compared to 
LLMs trained on general corpora. 
This finding underscores the importance of domain-specific adaptation and
highlights the potential of leveraging real-world clinical data 
to improve natural language processing solutions.

We begin in \Cref{sec:preliminaries} with the presentation of our chosen baseline pretrained models.
Then, in \Cref{sec:dataset} we describe the medical dataset and the work we have done to 
de-identify medical texts. 
Successively, we present the benchmark datasets, and in \Cref{sec:traineval} we present
our training setup, and the evaluation results.
We conclude with a final discussion.

\section{Preliminaries}
\label{sec:preliminaries}

The foundation of our work rests upon recent advances in pre-trained language models, 
specifically transformer-based encoders, which have demonstrated remarkable capabilities
in finding patterns in natural language. 
By using these models we can
transfer knowledge learned from extensive textual data, while using less resources.
This section details the
pre-trained encoder models utilized in this study, outlining their training data, architectural
characteristics, and key methodological distinctions. We employ three distinct 
language models: \nbbert~\cite{nbbert}, \norbert~\cite{modernbert}, and \modernbert~\cite{modernbert}, each contributing unique strengths
to our overall approach.

\vspace{3pt}
{\bfseries \nbbert}
~\cite{nbbert} was developed by the National Library of Norway\footnote{\url{https://www.nb.no/}}.
This model adopts the same architecture as the BERT Cased Multilingual model~\cite{DBLP}
by Google, a widely-used 
and well-known  encoder model with $356$ million parameters. However, unlike its multilingual counterpart, 
\nbbert is trained from scratch on a comprehensive collection of Norwegian text
spanning the last $200$ years. This training process utilizes a monolingual Norwegian
vocabulary, ensuring a deep representation of the nuances of both Bokmål and Nynorsk
written forms. The model was trained using standard language modeling objectives: 
masked language modeling (MLM) with a $15\%$ masking rate, and next sentence prediction
(NSP) with a maximum context length of $512$ tokens. It was trained on the Norwegian
Colossal Corpus~\cite{kummervoldetal}, a large dataset comprising approximately
$6.9$ billion words of Norwegian text. The intentional focus on Norwegian language data
and the inclusion of both major written standards position \nbbert as a strong baseline
for tasks requiring contextualized Norwegian language representations. 

\vspace{3pt}
{\bfseries \norbert.}
Developed by the Language Technology Group
at the University of Oslo, \norbert~\cite{norbert} presents a more extensively trained and multi-task oriented
model. While also based on a transformer architecture, \norbert distinguishes itself
through its training methodology, which incorporates a wider range of supervised tasks
alongside self-supervised learning. 
This model was tested on 
a suite of downstream tasks important for Norwegian natural language processing,
including Bokmål and Nynorsk standardization, lemmatization, dependency parsing, named
entity recognition, sentiment classification and analysis, question answering, and
machine translation between Bokmål and Nynorsk. 
Also this model handles a maximum context length of $512$ tokens  and has a size of $323$ million parameters.
The training data for \norbert is
significantly larger than that used for \nbbert, totaling approximately $25$ billion
words.  This dataset is a composite of several prominent Norwegian corpora, including
a Norwegian Wikipedia dump (October 2022), NBDigital, the Norwegian Colossal Corpus,
and the Norwegian portion of the mC4 corpus.  
Notably, \norbert employs the span masked language
modeling~\cite{spanbert} until 15\% input text is masked while training and
with a $512$ context length but omits the next sentence prediction objective
as it was shown to be unnecessary~\cite{liu2019} by later studies.

\vspace{3pt}
{\bfseries \modernbert.}
Finally, \modernbert~\cite{modernbert}, developed by answer.ai et al., represents a more recent
approach to model architecture design.
This model departs from the standard
BERT architecture with a novel design featuring the removal of bias terms and the implementation
of alternating attention and positional embeddings. \modernbert was trained with data 
on another scale compared to the other two models,
leveraging approximately $2$ trillion tokens sourced from diverse data repositories.
Moreover, thanks to the architectural improvements, this model can handle a maximum context length of 
$8192$ tokens with a size of $395$ million parameters.
The training methodology centers on masked language modeling with a relatively high
masking rate of $30\%$, again without the inclusion of a next-sentence prediction task.
The sheer scale of the training data and the architectural innovations implemented
in \modernbert 
enables the processing of longer texts and as a consequence
the capturing of more linguistic patterns.

These three models \nbbert, \norbert and \modernbert  provide a comprehensive
landscape of pre-trained encoder models suited for 
pretraining with the goal of further training these models with additional specialized data
to automate single jobs.

\section{Dataset}
\label{sec:dataset}

In this section, we describe the dataset used to pretrain our specialised clinical language models and 
the benchmark datasets used for evaluation the performance with respect to the base non-clinical pretrained models. 

\subsection{Clinical Dataset}
\subsubsection{Population}
The training dataset is based on all clinical relevant text documents from almost all patients of 
the hospitals of Helse Vest, which includes hospitals from Helse Bergen, 
Helse Fonna, Helse Førde and Helse Stavanger, 
in the years 2019, 2021 and 2022. We exclude data from 2020 because of concerns regarding a Covid-19 bias. 
We excluded patients with a confidential adress in the norwegian National Population Register and patients 
which wanted to be excluded. This added up to a non-significant number of excluded people 
(less than $0,05\%$) 
such that the dataset is viewed as highly representative for the hospitals population in Helse Vest.
We want to stress that we included all medical specialities (inluding mental health) 
and all age groups.

The data were exported from the electronical health journal DIPS, with the help of technical experts from 
DIPS and Helse Vest ICT. DIPS has over $1,100$ different document types, but not all included 
relevant information for this project. We included $715$ of them in our 
dataset based on their clinical relevance, i.e. we excluded documents with only adminstrational information 
and no medical relevance. 

In total, our dataset included $16,211,880$ unique documents with $89\%$ written in bokmål 
and $11\%$ in nynorsk. The documents included $456,063,804$ sentences and around $5.1$ billion tokens 
with a median number of tokens per document of around $203$. 
This dataset produced around $38$ million unique trainingpoints. 

\subsubsection{De-identification}
All documents were automatically de-identified before used in training. The de-identification 
application was developed for this project and de-identified the following categories: 
Names, locations, Health care units, Dates (including years), personal identification numbers, 
phone numbers and email-adresses. The categories were replaced by random but realistic values 
using the "hidden-in-plain-sight" methodology such that 
the name "Peter" could be replaced by "Karl" for example. The same word would be replaced by the same 
value throughout the same document.
All dates in one document were shifted by the same 
random number of days, which was between 365 and 1,825, into the future or past such that that chronological 
consistency was preserved throughout the document.

\begin{figure}
    \centering
            \includegraphics[totalheight=8cm]{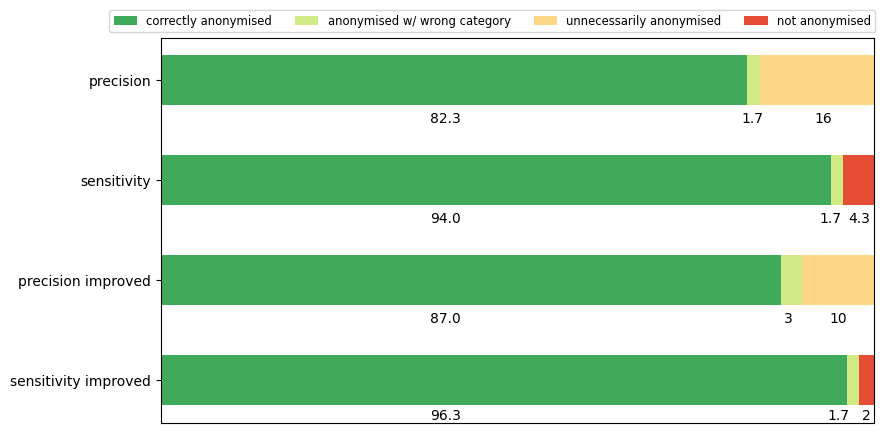}
        \caption{precision and recall in percentage.}
        \label{fig:evalanon}
\end{figure}

The de-identification application was evaluated on $1,000$ randomly choosen documents from 
a population of around $1$ million documents.
The initial evaluation results, depicted in  the first two columns of \Cref{fig:evalanon},  
showed that the de-identification application reduced privacy risks 
for the patients with an identification rate of $95.7\%$ of all sensitive information 
($94.0\%$ with the correct category) and $16\%$ unnecessary de-identifications of 
non-sensitive information. We found no critical errors in de-identification that would directly 
identify patients.  
Most detected errors affected the quality of the de-identified texts, but 
not the patient's privacy.

After the de-identification application was improved
we manage to reach 
an identification rate of $98.0\%$ of all sensitive information 
($96.3\%$ with the correct category) and only $10.0\%$ unnecessary de-identification of 
non-sensitive information, results depicted in the last two rows of \Cref{fig:evalanon}.

Finally, the project evaluated that the resulting de-identified 
texts are of good quality for training of the model.

\subsection{Evaluation Dataset}

To facilitate robust evaluation and development of medical language 
models, we utilize and present five distinct benchmark datasets, each designed to
address specific challenges within the domain.

\vspace{3pt}
{\bfseries \medMCQA.}
The first,
\medMCQA~\cite{pmlrv174pal22a}
is a large-scale,
Multiple-Choice Question Answering (MCQA) dataset in english constructed from real-world medical
entrance exam questions. This dataset is formulated as a task of selecting the correct
answer(s) from a set of candidate options.
It can be formulated as a multi-classification task where each question ($1$ to $4$)
corresponds to a single class.
\medMCQA comprises over $194.000$ high-quality MCQs
sourced from AIIMS and NEET PG entrance exams, providing comprehensive coverage of
$2,400$ healthcare topics across $21$ medical subjects. The questions within MedMCQA exhibit
an average token length of $12.77$, indicating a reasonable complexity, and are characterized
by high topical diversity, ensuring a broad assessment of model understanding. This
scale and breadth make MedMCQA a valuable resource for evaluating a model's capacity
to recall and apply medical knowledge in a realistic  setting.

We machine translated (using Google Translate) the dataset to bokmål. We are aware 
of the risk of errors in the translation, which could lead to non-sensical questions or answers. 
But the results of our evaluation~\ref{tab:res} lead to the conclusion that the 
translated dataset is of good quality.

\vspace{3pt}
{\bfseries \medMQP.}
Our second dataset, the Medical Question Pairs (MQP) Dataset~\cite{mccreery2020} in english, 
focuses on the nuanced
task of semantic similarity in patient-generated questions.  Medical professionals were tasked to
generate question pairs from a randomly sampled set of $1,524$ patient questions
obtained from a public HealthTap crawl. For each original question, labelers were instructed
to create both a positive pair, that is a question rephrased to maintain the same intent but
with altered syntax and minor medical detail changes, and a negative pair, a related
question for which the original question's answer would be incorrect or irrelevant.
Specifically, the positive pair instruction encouraged significant restructuring and
re-wording (e.g., transforming “I'm a $22$-year-old female” to “My 26-year-old daughter”),
while the negative pair instruction leveraged similar keywords to create deceptively
similar but semantically distinct questions. This intentional framing of the task is
critical; it mitigates the risk of superficial similarity-based learning by requiring
models to understand underlying intent rather than relying on lexical overlap.  The
resulting dataset contains $4,567$ unique questions, effectively doubling the dataset
size through the creation of positive and negative pairs, and presents a challenging
benchmark for semantic understanding.

We machine translated (using Google Translate) the dataset to bokmål. We are aware 
of the risk of errors in the translation, which could lead to non-sensical question-pairs. 
But the results of our evaluation~\ref{tab:res} lead to the conclusion that the 
translated dataset is of good quality.

\vspace{3pt}
{\bfseries \medner.}
Finally, we use the Nor-DeID-SynthData~\cite{pmlrv233lund24a}, 
a synthetic dataset of Norwegian discharge
summaries. Originally created for benchmarking de-identification methods, this dataset
is a valuable resource for training and evaluating Named Entity Recognition (NER) models.
The dataset is designed to facilitate the
identification of sensitive patient information, specifically focusing on entities
such as first and last names, age, phone numbers, locations, healthcare units, social
security numbers, and dates/years.  With a total of $1,000$ examples, Nor-DeID-SynthData
allows for the training of models capable of extracting and redacting protected health
information from clinical text written in Norwegian.  

\vspace{3pt}
{\bfseries \AISMEC.}
Our fourth dataset is the first non-synthetic, consisting only of original norwegian texts 
from the Helse Vest health care region and 
defines a classical binary classification task. In detail, Artificial Intelligence Support 
in Medical Emergency Calls (\AISMEC) is a research 
project aimed at developing decision support using artificial intelligence (AI) for 
operators at the medical emergency number (113). By analyzing what is said during 
emergency calls and combining it with the patient’s hospital history, medication use, and 
blood test results, the AI model can estimate the risk of stroke. 
The dataset used was balanced and consisted of in total ca. $1,130$ transcripts of patients 
calling 113, i.e.around $565$ having strokes and $565$ not having strokes.

\vspace{3pt}
{\bfseries \Synergi.}
Our fifth dataset is a non-synthetic, consisting only of original but de-identified norwegian 
texts from Helse Bergens quality and EHS management software Synergi and defines a 
NER task. The dataset is part of the Helse Vest innovation project "Artificial Intelligence 
Support in Medical Emergency Calls", which aimed for identifying all mentioned drugs in a 
given text, independent of the official or a non-official drug name was used. These texts 
can be written by anybody working in the hospital such that it is a uniquely diverse dataset 
with respect to styles and medical slangs used. 
The dataset used consists of $1,992$ texts and in total $2,706$ drug labels. The labeling was done 
manually by qualified health care professionals and was iteratively improved during the project. 
This result is a dataset of high quality.

Across the four datasets \medMCQA, 
\medMQP, \medner, and \AISMEC, we maintain a consistent data split of $75\%$ for training, 
$10\%$ for validation, and $15\%$ for testing. For \Synergi, the data split was $80\%$ for training, 
$10\%$ for validation, and $10\%$ for testing.

\section{Training and Evaluation}
\label{sec:traineval}

In this section, we give a general overview of our computing resources and list the hyperparameters chosen
that gave the best results while validating such models.
Then, in the next subsection we present the evaluation results.

\subsection{Training}
All models were trained on a high-performance computing machine equipped with four
NVIDIA L40S-48C GPUs, each possessing 48GB of memory, coupled with eight INTEL(R) Xeon
(R) Gold 6548N CPUs and 62GB of system RAM. The training corpus comprised approximately
5 billion tokens of text data. Due to the substantial size of the dataset and the memory
demands of transformer-based models, gradient accumulation was employed across all
training runs to effectively simulate larger batch sizes than would otherwise be impossible
to achieve
within the constraints of our available GPU memory. The training process for each
model required approximately one week to complete.
We used the pytorch library~\cite{paszke2019} 
and the huggingface library suite~\cite{wolf2020}: 
accelerate, datasets, evaluate etc. for multi-gpu model training and testing.
The pretraining procedure lasted only for $1$ epoch to ensure minimal overfitting.

\vspace{3pt}
{\bfseries \nbbert}
was trained using the AdamW optimizer~\cite{kingma2017} 
with a learning rate of
$2e-5$ and a cosine learning rate scheduler to facilitate stable convergence. A weight
decay of $0.01$ was applied as a regularization technique to prevent overfitting.  
Training
utilized a batch size of $64$ and a maximum context length of $512$ tokens. 
The pre-training
task consisted of standard masked language modeling (MLM), wherein $20\%$ 
of the input
tokens were randomly masked and predicted by the model, alongside the next sentence
prediction (NSP) objective. 

\vspace{3pt}
{\bfseries \norbert}
was trained using the AdamW optimizer~\cite{loshchilov2019}, 
with a learning rate of
$1e-5$ and a cosine learning rate scheduler, and a weight decay of $0.01$.  
A batch size
of $64$ was used, and the maximum context length was set to $512$ tokens, consistent with
the \nbbert training setup. Notably, \norbert's training deviated from the standard
MLM and NSP approach. Instead, we implemented span masked language modeling, mirroring
the methodology employed in the original \norbert publication.  This technique masks
contiguous spans of text, promoting a stronger focus on contextual understanding within
longer phrases, and we omitted the next sentence prediction task.

\vspace{3pt}
Finally,
{\bfseries \modernbert}
was trained using the StableAdamW 
optimizer~\cite{wortsman2023}, a variant designed for improved
training stability, with a fixed learning rate of $1e-5$ and without the use of a learning
rate scheduler or weight decay.  
A larger batch size of $256$ was utilized for ModernBERT's training. 
The pre-training task was based on masked language modeling; however,
a dynamic masking strategy was adopted. Initially, $15\%$ of the input tokens were masked
during the first $5\%$ of training iterations.  
This percentage was then gradually increased to $30\%$ over the remaining 
$95\%$ of the dataset. 
This progressive masking approach aimed to simulate the training of the baseline 
pretrained \modernbert model.

\subsection{Evaluation}

To rigorously assess the performance of our proposed models, we split our dataset 
into
three distinct subsets: $75\%$ for training, $10\%$ for evaluation, and $15\%$ for final, independent
testing.  
As opposed to the work done in \modernbert~\cite{modernbert}(Appendix A.4), 
we did not employ averaging methods from the last best performing checkpoints for 
finding the best final model after training, 
opting instead to select the single model achieving the best performance on the evaluation
dataset. 
While averaging checkpoints could potentially improve the performance
we chose to carry a direct comparison with respect to the other models (our finetuned or baselines).
The final reported results are then based solely on the
performance of these 'best` models on the independent test set. 

Following training and selection, we evaluated the performance of our
three newly trained models alongside the three baseline models. 
The results are summarized in \Cref{tab:res}. 
The table presents the F1-score
for each model on each of the three syntethic datasets. For the real-world evaluations in the \AISMEC and 
\Synergi projects we just used one model category because of resource constraints.

\renewcommand{\arraystretch}{1.3}
\begin{table}[h!]
\centering
\caption{F1-score on Test Datasets.}
\bigskip
\begin{tabular}{l|c|c|c|c|c}
\hline
\textbf{Model Name} & \textbf{\medMCQA} & \textbf{\medMQP} & \textbf{\medner}& \textbf{\AISMEC}& \textbf{\Synergi} \\
\hline
\knbbert & {\bfseries 39.17} & {\bfseries 89.64}  & 89.75 &---& {\bfseries 94.58} \\
\knorbert & 34.15  & 70.29 & 88.67 &---&---\\
\kmodernbert & 36.7 & 70.37 & {\bfseries 90.49}& {\bfseries 82.14} &--- \\
\hline
\nbbert & 28.44 & 84.12  & 88.19 &---  & 88.02 \\
\norbert & 32.51  & 64.04 & 88.49&---&---\\
\modernbert & 31.55 & 67.92 & 88.8& 79.52 &--- \\
\hline
\end{tabular}
\label{tab:res}
\end{table}

All clinical models perform better than their respective baseline models.
\knbbert seems to be the preferred model in case the input text is short due to its limited context length.
Otherwise, \kmodernbert seems to be the best alternative, especially in tasks requireing identification
of parts of texts relevant to the task to be automated, like NER.

To further assess the generalizability of our clinical models, we performed validation
testing using several independent datasets. While the sensitive nature of this data
prevents public sharing, these experiments 
consistently indicated robust performance, reinforcing the findings presented within
this work and suggesting a degree of transferability of the learned representations.

A notable observation during model training was the significantly faster convergence
rate exhibited by our clinical models compared to non-clinical baselines. 
Specifically,
in the context of a Named Entity Recognition (NER) task focused on identifying medical
entities within text, clinical models consistently achieved optimal checkpoint performance
(as determined by evaluation on a held-out dataset) within just 1-2 training epochs.
In contrast, baseline models required approximately 8-10 epochs to reach comparable
levels of performance. This accelerated learning suggests that the pre-training process
on clinical text effectively equips these models with prior knowledge relevant to the
task, allowing for more efficient adaptation and reduced training time.

\section{Conclusion}

This paper presented \suite, 
a suite of three BERT-based encoder models,
based on \nbbert, \norbert, and \modernbert,
pre-trained for enhanced performance
on Norwegian clinical language processing tasks. 
Recognizing the limitations of general-purpose
language models when applied to the nuanced and specialized terminology of healthcare,
we undertook a domain-specific pre-training strategy using a large, curated
corpus of real-world, de-identified Norwegian clinical texts. 
The resulting models
demonstrate consistent and substantial improvements over their baseline, 
non-clinical
counterparts when evaluated on established Norwegian clinical benchmark datasets.
This development represents a step towards 
more accurate and efficient  solutions
inherent in real-world clinical workflows. 
Moreover, these
findings underscore the critical importance of domain adaptation in achieving optimal results for NLP applications
within the clinical domain. 

\section{Acknowledgment}
This project could not have happend without the dedicated support of our employee Helse Vest ICT. 
Helse Vest ICT made a strategic decision to support this AI project to improve 
its own AI compute and AI development capabilities and add clinical specific 
AI tools to their services which already support several new AI projects inside 
Helse Vest. 

Data is the most important ingridient in model training and we would like to thank 
Helse Bergen, Helse Fonna, Helse Førde and Helse Stavanger for supporting the project 
with their data. Additionally, all of them provided privacy and legal experts 
to secure a patient and privacy friendly project.

DIPS was also a supportive partner from the beginning to the end and added epxertise 
knowledge which supported data export and data understanding to the project.

We would like to thank Styringsdata (Helse Vest ICT) for their help with building an efficient data export 
pipeline and supporting a secure data management strategy in the project. 

We thank Server og Lagring (Helse Vest ICT) for providing norwegians best GPU infrastructure inside 
the healthcare sector. The possibillity to use up to four GPUs simultaneously 
in a secure environment reduced the trainingperiode from planned three month to 
approximately one week.

We would like to thank the independent projects \AISMEC and \Synergi for 
giving us the opportunities to evaluate our models. These evaluations gave us 
significant inside in the real world implications of our work.



\nocite{*}
\printbibliography

\end{document}